\documentclass{llncs}

\usepackage{graphicx}
\usepackage{amsmath,amsfonts}
\usepackage{bbm} 
\usepackage{booktabs} 

\begin{document}
\title{Object proposal generation applying the distance dependent Chinese restaurant process}
\author{Mikko Lauri \and Simone Frintrop}
\institute{University of Hamburg\\ Department of Informatics\\ \email{ \{lauri,frintrop\}@informatik.uni-hamburg.de}}
\maketitle

\begin{abstract}
In application domains such as robotics, it is useful to represent the uncertainty related to the robot's belief about the state of its environment.
Algorithms that only yield a single ``best guess'' as a result are not sufficient.
In this paper, we propose object proposal generation based on non-parametric Bayesian inference that allows quantification of the likelihood of the proposals.
We apply Markov chain Monte Carlo to draw samples of image segmentations via the distance dependent Chinese restaurant process.
Our method achieves state-of-the-art performance on an indoor object discovery data set, while additionally providing a likelihood term for each proposal.
We show that the likelihood term can effectively be used to rank proposals according to their quality.
\end{abstract}

\section{Introduction} 
\label{sec:introduction}
Image data in robotics is subject to uncertainty, e.g., due to robot motion, or variations in lighting.
To account for the uncertainty, it is not sufficient to apply deterministic algorithms that produce a single answer to a computer vision task.
Rather, we are interested in the full Bayesian posterior probability distribution related to the task; e.g., given the input image data, how likely is it that a particular image segment corresponds to a real object?
The posterior distribution enables quantitatively answering queries on relevant tasks which helps in decision making.
For example, the robot more likely succeeds in a grasping action targeting an object proposal with a high probability of corresponding to an actual object~\cite{vanHoof2014,Pajarinen2015b}.

\begin{figure}[ht]
	\centering
	\includegraphics[width=\columnwidth]{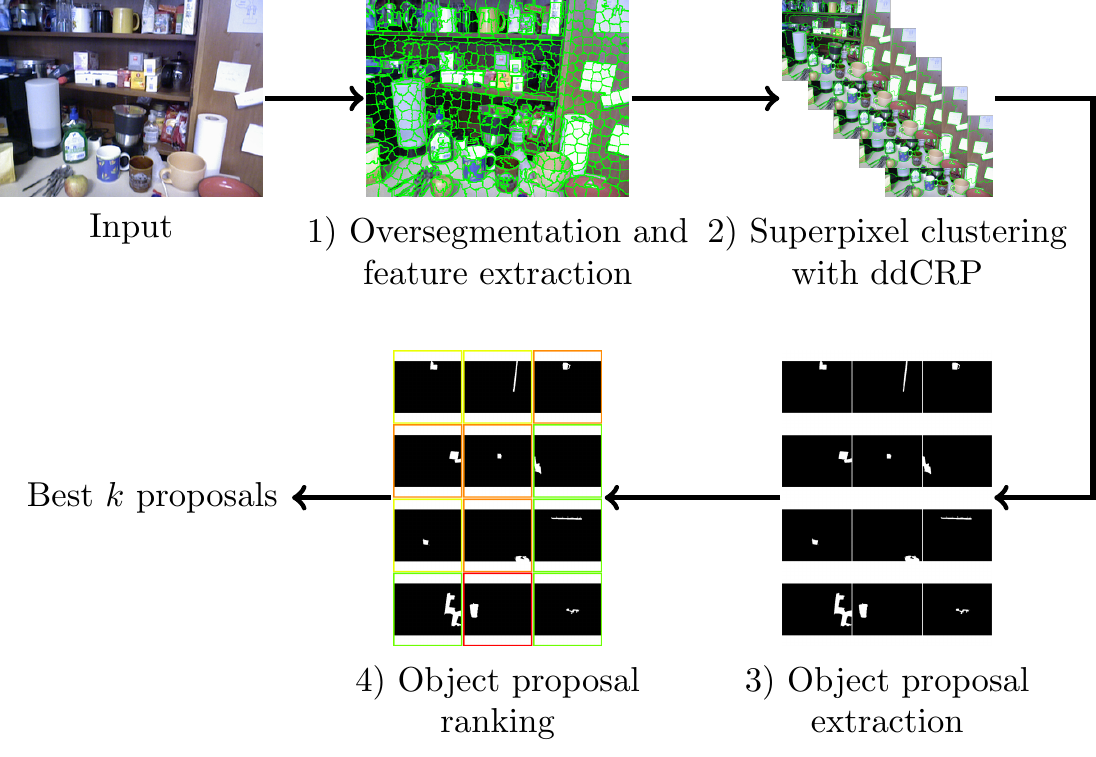}
	\caption{Overview of the object discovery approach. Superpixels in an initial oversegmentation (1) are grouped applying the distance dependent Chinese restaurant process (ddCRP) (2). Multiple segmentation samples are drawn from the ddCRP posterior distribution. Object proposals are extracted from the set of segmentation samples (3), and ranked according to how likely they correspond to an object (4).}
	\label{fig:overview}
\end{figure}

In this paper, we propose a method for object discovery based on the distance dependent Chinese restaurant process (ddCRP). 
In contrast to other approaches, we do not combine superpixels deterministically to generate object proposals, but instead place a ddCRP prior on clusters of superpixels, and then draw samples from the posterior given image data to generate proposals.
This firstly increases the diversity of object proposals, and secondly enables calculation of a likelihood term for each proposal.
We show that the likelihood term may be used to rank proposals according to their quality.
Additionally, the likelihood term might be exploited by a mobile robot to plan its actions.

An overview of our approach is shown in Fig.~\ref{fig:overview}.
We begin with a superpixel oversegmentation of the input image, and then place a ddCRP prior on clusters of superpixels.
The ddCRP hyperparameters are selected to encourage object proposal generation: clusters of superpixels with high internal similarity and external dissimilarity are preferred.
We apply Markov chain Monte Carlo (MCMC) to draw samples of the posterior distribution on clusterings of superpixels.
We extract all unique clusters which form our set of object proposals.
We rank the object proposals according to the Gestalt principles of human object perception~\cite{Wagemans2012}.
We propose to include the likelihood term, i.e., how often each proposal appears in the set of samples, as part of the ranking, and show that this effectively improves the quality of the proposals.

The paper is organized as follows.
Section~\ref{sec:related_work} reviews related work and states our contribution w.r.t. the state-of-the-art.
In Sections~\ref{sec:the_distance_dependent_chinese_restaurant_process}-\ref{sec:gestalt_principles_for_object_discovery}, we present in detail the steps involved in the overall process shown in Fig.~\ref{fig:overview}.
Section~\ref{sec:evaluation} describes an experimental evaluation of our approach.
 Section~\ref{sec:conclusion} concludes the paper.


\section{Related work} 
\label{sec:related_work}
Object discovery methods include window-scoring methods (e.g.~\cite{Alexe2012}) that slide a window over the image which is evaluated for its objectness, and segment-grouping methods (e.g.~\cite{Manen2013}), that start with an oversegmentation of the image and group these segments to obtain object proposals.
Segment-grouping methods have the advantage of delivering object contours instead of only bounding boxes, which is especially important in applications such as robotics where the object of interest might have to be manipulated. 
We concentrate here on the segment-grouping approach.

The segment-grouping approaches often start from an oversegmentation of the image into superpixels that are both spatially coherent and homogeneous with respect to desired criteria, e.g., texture or color.
Object proposals are then generated by combining several superpixels together.
For an overview of the various combination strategies we refer the reader to~\cite{Hosang2016}.

Although some segment-grouping approaches such as e.g.~\cite{Manen2013} apply random sampling to generate object proposals, it is often not possible to estimate a likelihood value for a particular combination of superpixels, nor is it intuitively clear what the overall probability distribution over image segments is that is applied in the sampling.
However, both these properties are useful in application domains such as robotics, where decisions are made based on the observed image data, see, e.g.,~\cite{vanHoof2014,Pajarinen2015b}.
To address these limitations, we consider non-parametric Bayesian methods for superpixel clustering.
Such methods have been previously applied to image segmentation with the aim of replicating human segmentation of images.
For example, \cite{Ghosh2011} applies the distance dependent Chinese restaurant process (ddCRP) and \cite{Nakamura2012} proposes a hierarchical Dirichlet process Markov random field for the segmentation task.
In~\cite{Ghosh2012}, multiple segmentation hypotheses are produced applying the spatially dependent Pitman-Yor process.
Recent work applies a Poisson process with segment shape priors for segmentation~\cite{Ghanta2016}.

In our work, similarly to~\cite{Ghosh2011}, we apply Markov chain Monte Carlo (MCMC) sampling from a ddCRP posterior to generate clusters of superpixels.
However, in contrast to earlier work our main aim is object discovery.
We tune our method especially towards this aim by setting the model hyperparameters to produce clusters of superpixels that have a strong link to human object perception as described by the Gestalt principles of human object perception~\cite{Wagemans2012}.


\section{The distance dependent Chinese restaurant process} 
\label{sec:the_distance_dependent_chinese_restaurant_process}
We first oversegment the input image into superpixels (step 1 in Fig.~\ref{fig:overview}).
For each superpixel, we compute a feature vector $x_i$ that we define later.
We generate object proposals by grouping superpixels together applying the distance dependent Chinese restaurant process (ddCRP)~\cite{Blei2011}, a distribution over partitions.

\begin{figure}[t]
	\centering
	\includegraphics{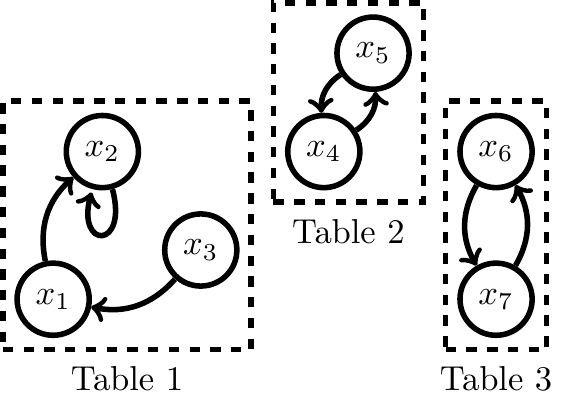}
	\caption{The distance dependent Chinese restaurant process. Customers corresponding to superpixels in the input image are denoted by the nodes $x_i$. The links between customers induce a table assignment which corresponds to a segmentation of the image.}
	\label{fig:ddcrp}
\end{figure}

The ddCRP is illustrated by an analogy where data points correspond to customers in a restaurant.
Every customer links to another customer with whom they will sit at the same table.
A partitioning is induced by this set of customer links: any two customers $i$ and $j$ are seated at the same table if $i$ can be reached from $j$ traversing the links between customers (regardless of link direction).
Applied to object proposal generation, the image is the restaurant, the customers are superpixels, and the assignment of customers to tables corresponds to a segmentation of the image, with each table forming one object proposal -- see Fig.~\ref{fig:ddcrp} for an illustration.

In the ddCRP, the probability that a customer links to another is proportional to the distance between customers.
Let $c_i$ denote the index of the customer linked to by customer $i$, $d_{ij}$ the distance between customers $i$ and $j$, and $D$ the set of all such distances.
The customer links are drawn conditioned on the distances,
\begin{equation}
	p(c_i = j \mid D, f, \alpha) \propto \begin{cases} \alpha & \text{if }j = i \\ f(d_{ij}) & \text{if }j\neq i\end{cases},
	\label{eq:ddcrp_prior}
\end{equation}
where $\alpha$ is a parameter defining the likelihood of self-links, and $f:[0,\infty)\to\mathbb{R}^+$ is a decay function that relates the distances between customers to the likelihood of them connecting to each other.
We require $f$ to be non-increasing and $f(\infty) = 0$.

We next define the posterior over customer links.
Let $\boldsymbol{x} = x_{1:N}$ denote the collection of all data points.
Denote by $\boldsymbol{c} = c_{1:N}$ the vector of customer links, and by $\boldsymbol{z}(\boldsymbol{c})$ the corresponding vector of assignments of customers to tables.
Denote by $K \equiv K(\boldsymbol{c})$ the number of tables corresponding to link assignment $\boldsymbol{c}$.
Furthermore, write $\boldsymbol{z}^k(\boldsymbol{c})$ for the set of all customers $i$ that are assigned to table $k\in \{1, \ldots K\}$.
For each table $k$, we assume that the data $x_i$, $i \in \boldsymbol{z}^k(\boldsymbol{c})$, is generated from $p(\cdot \mid \theta_k)$.
The parameter $\theta_k$ is assumed to be drawn from a base measure $G_0$, which may be considered a prior on $\theta$.
Thus, the posterior is
\begin{equation}
	p(\boldsymbol{c} \mid \boldsymbol{x}, D, f, \alpha, G_0) \propto \left(\prod\limits_{i=1}^N p(c_i\mid D, f, \alpha) \right) p(\boldsymbol{x} \mid \boldsymbol{z}(\boldsymbol{c}), G_0).
	\label{eq:ddcrp_posterior}
\end{equation}
The first term on the right hand side above is the ddCRP prior, and the second likelihood term is conditionally independent between the tables $k$:
\begin{equation}
	 p(\boldsymbol{x} \mid \boldsymbol{z}(\boldsymbol{c}), G_0) = \prod\limits_{k=1}^K p(\boldsymbol{x}_{\boldsymbol{z}^k(\boldsymbol{c})} \mid G_0),
	\label{eq:ddcrp_likelihood}
\end{equation}
where $\boldsymbol{x}_{\boldsymbol{z}^k(\boldsymbol{c})}$ denotes the collection of data points in table $k$ under link configuration $\boldsymbol{c}$.
As the ddCRP places a prior on a combinatorial number of possible image segmentations, computing the posterior is not tractable.
Instead, we apply Markov chain Monte Carlo (MCMC)~\cite[Sect.~24.2]{Murphy2012} to sample from the posterior given the model hyperparameters $\boldsymbol{\eta} = \{D, f, \alpha, G_0\}$.

\paragraph{Sampling from the ddCRP posterior:} 
\label{ssub:gibbs_sampler_for_the_ddcrp}
Sampling from the ddCRP corresponds to step 2 of Fig.~\ref{fig:overview}, and each individual sample corresponds to a segmentation of the input image - see Fig.~\ref{fig:seg_example}, left, for an example.
We apply Gibbs sampling, a MCMC algorithm for drawing samples from high-dimensional probability density functions, introduced for the ddCRP in~\cite{Blei2011}.
The idea is to sample each variable sequentially, conditioned on the values of all other variables in the distribution.
Denote by $\boldsymbol{c}_{-i}$ the vector of link assignments excluding $c_i$.
We sequentially sample a new link assignment $c_i^*$ for each customer $i$ conditioned on $\boldsymbol{c}_{-i}$ via
\begin{equation}
	p( c_i^{*} \mid \boldsymbol{c}_{-i}, \boldsymbol{x}, \boldsymbol{\eta}) \propto p(c_i^* \mid D, f, \alpha) p(\boldsymbol{x} \mid \boldsymbol{z}(\boldsymbol{c}_{-i} \cup c_i^*), G_0).
	\label{eq:ddcrp_gibbs}
\end{equation}
The first right hand side term is the ddCRP prior of Eq.~\eqref{eq:ddcrp_prior}, and the second term is the marginal likelihood of the data under the partition $\boldsymbol{z}(\boldsymbol{c}_{-i} \cup c_i^*)$.
The current link $c_i$ is first removed from the customer graph which may either cause no change in the table configuration, or split a table (c.f. Fig.~\ref{fig:ddcrp}).
Then, reasoning about the effect that a potential new link $c_i^*$ would have on the table configuration, it can be shown that~\cite{Blei2011}
\begin{equation}
	p( c_i^{*} \mid \boldsymbol{c}_{-i}, \boldsymbol{x}, \boldsymbol{\eta}) \propto 
	\begin{cases}
		\alpha & \text{if } c_i^* = i\\
		f(d_{ij}) & \text{if } c_i^*=j \text{ does not join two tables}\\
		f(d_{ij}) L(\boldsymbol{x}, \boldsymbol{z}, G_0) & \text{if } c_i^* = j \text{ joins tables } k \text{ and } l,
	\end{cases}
	\label{eq:ddcrp_gibbs}
\end{equation}
where
\begin{equation}
	L(\boldsymbol{x}, \boldsymbol{z}, G_0) = \frac{p(\boldsymbol{x}_{ \boldsymbol{z}^k(\boldsymbol{c}_{-i}) \cup \boldsymbol{z}^l(\boldsymbol{c}_{-i}) } \mid G_0) }{ p(\boldsymbol{x}_{ \boldsymbol{z}^k(\boldsymbol{c}_{-i})}\mid G_0) p(\boldsymbol{x}_{ \boldsymbol{z}^l(\boldsymbol{c}_{-i})}\mid G_0) }.
	\label{eq:likelihood_ratio}
\end{equation}
The terms in the nominator and denominator can be computed via
\begin{equation}
	p(\boldsymbol{x}_{\boldsymbol{z}^k(\boldsymbol{c})} \mid G_0) = \int \left( \prod\limits_{i \in \boldsymbol{z}^k(\boldsymbol{c})} p(x_i \mid \theta)\right) p(\theta \mid G_0) d\theta.
	\label{eq:cluster_likelihood}
\end{equation}
Recall that we interpret the base measure $G_0$ as a prior over the parameters: $G_0 \equiv p(\theta)$.
If $p(\theta)$ and $p(x\mid \theta)$ form a conjugate pair, the integral is usually straightforward to compute.


\section{Object proposal generation and likelihood estimation} 
\label{sec:object_proposal_generation}
We extract a set of object proposals (step 3 in Fig.~\ref{fig:overview}) from samples drawn from the ddCRP posterior.
Furthermore, we associate with each proposal an estimate of its likelihood of occurrence.
As proposals are clusters of superpixels, we use here notation $s_i$ to refer to superpixels instead of their feature vectors $x_i$.

To sample a customer assignment $\boldsymbol{c}$ from the ddCRP posterior, we draw a sample from Eq.~\eqref{eq:ddcrp_gibbs} for each $i=1,\ldots,N$.
Denote by $\boldsymbol{c}_j$ the $j$th sample, and by $K_j \equiv K(\boldsymbol{c}_j)$ the number of tables in the corresponding table assignment.
We can view $\boldsymbol{c}_j$ as a segmentation of the input image, $\bigcup\limits_{k=1}^{K_j} S_{j,k}$, where $S_{j,k} = \{ s_i \mid i \in \boldsymbol{z}^k(\boldsymbol{c}_j) \}$ is the set of superpixels assigned to table $k$ by $\boldsymbol{c}_j$.
E.g., in Fig.~\ref{fig:ddcrp}, we would have $S_{j,1} = \{s_1, s_2, s_3\}$, $S_{j,2} = \{s_4, s_5\}$, and $S_{j,3} = \{s_6, s_7 \}$.

We sample $M$ customer assignments $\boldsymbol{c}_j$, $j=1, \ldots, M$, and write $S_j = \{ S_{j,1}, S_{j,2}, \ldots, S_{j, K_j} \}$ as the set of segments in the $j$th customer assignment. 
E.g., for the case of Fig.~\ref{fig:ddcrp}, we have $S_j = \{ S_{j,1}, S_{j,2}, S_{j,3}\}$ $=\{ \{s_1,s_2,s_3\}$, $\{s_4,s_5\}$, $\{s_6,s_7\} \}$.
The set $O$ of object proposals is obtained by keeping all unique segments observed among the sampled customer assignments: $O = \bigcup\limits_{j=1}^{M} S_j$.

\begin{figure}[t]
	\centering
	\includegraphics[width=\columnwidth]{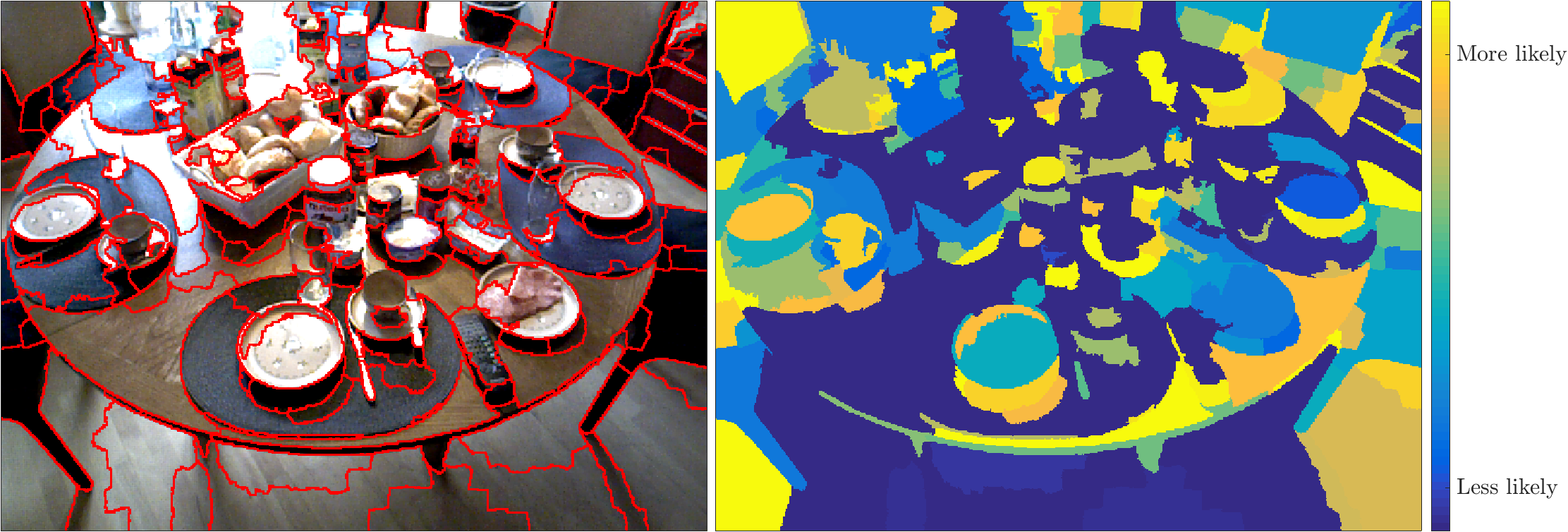}
	\caption{Left: an example of a segmentation result from the ddCRP. Each segment is a proposal $o$. Right: The corresponding proposal likelihood estimates $P(o)$.}
	\label{fig:seg_example}
\end{figure}

Each proposal $o\in O$ appears in at least one and in at most $M$ of the assignments $S_j$, $j=1,\ldots,M$.
We estimate the likelihood of each proposal by
\begin{equation}
	P(o) = \left. \left[ \sum\limits_{j=1}^{M} \mathbbm{1}\left( o \in S_j \right) \right] \middle/ \left[ \sum\limits_{j=1}^{M}|S_j| \right] \right.,
	\label{eq:proposal_likelihood}
\end{equation}
where $\mathbbm{1}(A)$ is an indicator function for event $A$, and $|\cdot|$ denotes set cardinality.
Fig.~\ref{fig:seg_example} illustrates the likelihood values for the proposals.

\section{Gestalt principles for object discovery} 
\label{sec:gestalt_principles_for_object_discovery}
We select the hyperparameters $\boldsymbol{\eta} = \{D, f, \alpha, G_0\}$ to promote two important principles: objects tend to have \emph{internal consistency} while also exhibiting \emph{contrast} against their background.
This ensures that the proposal set $O$ contains segments that are likely to correspond to objects.
As $O$ contains segments from all parts of the image, there are certainly also segments that belong to the background and contain no objects.
To mitigate this drawback, we rank the proposals in $O$ and output them in a best-first order.
For ranking, we calculate a set of scores from the proposals based on properties such as convexity and symmetry, that have also been shown to have a strong connection to object perception~\cite{Wagemans2012}.
Next, we describe the superpixel feature extraction, the selection of the ddCRP hyperparameters, and the ranking of object proposals (step 4 of Fig.~\ref{fig:overview}).

\paragraph{Feature extraction:} 
\label{ssub:feature_extraction}
We compute three feature maps from the input image as in: the grayscale intensity $I$, and the red-green and blue-yellow color contrast maps $RG$ and $BY$, respectively.
The feature vector $x_i$ for superpixel $i$ is
\begin{equation}
x_i = 
\begin{bmatrix}
	x_{i,I} &
	x_{i,RG} &
	x_{i,BY} &
	x_{i,avg}
\end{bmatrix}^\mathrm{T},
\end{equation}
where $x_{i,I}$, $x_{i,RG}$, and $x_{i,BY}$ are the 16-bin normalized histograms of the intensity, red-green, and blue-yellow contrast maps, respectively, and $x_{i,avg}$ is the average RGB color value in the superpixel.


\paragraph{Hyperparameter selection:} 
\label{ssub:hyperparameter_selection}
We incorporate contrast and consistency via the distance function $d$ and the base measure $G_0$, respectively.
The distance function $d$ and the decay function $f$ determine how likely it is to link two data points.
We impose a condition that only superpixels that share a border may be directly linked together.
Also, superpixels with similar contrast features should be more likely to be linked.
We define our distance function as
\begin{equation}
	d(i,j) = 
	\begin{cases}
		\infty & \text{if } s_i \text{ and } s_j \text{ are not adjacent}\\
		\sum\limits_{n\in \{I, RG, BY\}} w_n \cdot v(x_{i,n}, x_{j,n}) & \text{otherwise}
	\end{cases},
	\label{eq:distance}
\end{equation}
where $v(x, y) = \frac{1}{2} ||x-y||_1$ is the total variation distance, and $w_n$ is a weight for feature $n \in \{I, RG, BY\}$, s.t. $\sum_n w_n = 1$.
The distance function $d$ has values in the range $[0,1]$, or the value $\infty$. 
The weights $w_n$ may be tuned to emphasize certain types of contrasts, but in our experiments we set all to $1/3$.
We set an exponential decay function $f(d) = \exp(-d/a)$, where $a > 0$ is a design hyperparameter, to make it more likely to link to similar superpixels.

We encourage internal consistency in the segments by setting the base measure $G_0$.
For the likelihood terms in Eq.~\eqref{eq:cluster_likelihood}, we only consider the average RGB color feature $x_{i,avg}$ of the superpixels\footnote{The other elements of the feature vector are considered via the distance function $d$.}, which is a 3-dimensional vector.
We set a multivariate Gaussian cluster likelihood model $p(x_{i,avg} \mid \theta) = N(x_{i,avg}; \mu, \Sigma)$.
The model parameters are $\theta = \{\mu, \Sigma\}$, where $\mu$ and $\Sigma$ are the mean vector and covariance matrix, respectively.
We apply the Normal-inverse-Wishart distribution as a conjugate prior~\cite[Sect.~4.6.3]{Murphy2012}, i.e. $p(\theta \mid G_0) = NIW(\theta \mid m_0, \kappa_0, v_0, S_0) = N(\mu\mid m_0, \frac{1}{\kappa_0}\Sigma)\cdot IW(\Sigma\mid S_0, v_0)$.
Here, $m_0$, $\kappa_0$, indicate our prior mean for $\mu$ and how strongly we believe in this prior, respectively, and $S_0$ is proportional to the prior mean for $\Sigma$ and $v_0$ indicates the strength of this prior.
With this choice, adjacent superpixels with similar average RGB colors have a high likelihood of belonging to the same table in the ddCRP.


\paragraph{Object proposal ranking:} 
\label{sub:object_proposal_ranking}
Similarly as in~\cite{Werner2015}, for each object proposal $o\in O$, we compute the following Gestalt measures that have been shown to have a relation to human object perception~\cite{Wagemans2012}:
\begin{itemize}
	\item symmetry, calculated by measuring the overlaps $l_1$ and $l_2$ between the object proposal $o$ and its mirror images along both of its principal axes, i.e., eigenvectors of its scatter matrix. We use the symmetry measures $\frac{\lambda_1 l_1 + \lambda_2 l_2}{\lambda_1 + \lambda_2}$ and $\max \{ l_1, l_2\}$, where $\lambda_i$ are the eigenvalues of the scatter matrix,
	\item solidity, the ratio of the area of the convex hull of $o$ to the area of $o$ itself,
	\item convexity, the ratio of the proposal's boundary length and the boundary length of its convex hull,
	\item compactness, the ratio of the area of $o$ to the squared distance around the boundary of $o$, i.e., its perimeter,
	\item eccentricity, the ratio of the distance between the foci of the ellipse encompassing $o$ and its major axis length, and
	\item centroid distance, the average distance from the centroid of the proposal to its boundary.
\end{itemize}
As in~\cite{Werner2015}, we apply the first sequence of the KOD dataset~\cite{Horbert2015} to train a support vector machine (SVM) regression model~\cite[Sect.~14.5]{Murphy2012} from the Gestalt measures of a proposal $o$ to the intersection-over-union (IoU) of $o$ with the ground truth objects.

Applying the SVM, we can predict a score $s(o)$ for any object detection proposal in $O$.
The proposals with the highest score are deemed most likely to correspond to an actual object.
We propose a weighted variant of this score taking into account the likelihood (Eq.~\eqref{eq:proposal_likelihood}):
\begin{equation}
	s_w(o) = P(o)s(o).
\end{equation}
The rationale for this definition is that we would like to give higher priority to object proposals that 1) have a high score $s(o)$ and 2) appear often in the segmentations, indicating robustness with respect to internal consistency and external contrasts as defined via our model hyperparameters.
For example in Fig.~\ref{fig:seg_example}, the scores of proposals with high $P(o)$, i.e., proposals that appear in many samples from the ddCRP, are given higher priority.

As an optional step, we add non-maxima suppression (NMS) for duplicate removal: iterating over all object proposals $o$ in descending order of score, all lower ranked proposals with an IoU value greater than 0.5 with $o$ are pruned.



\section{Evaluation} 
\label{sec:evaluation}
We evaluate our object proposal generation method on the Kitchen Object Discovery (KOD) dataset~\cite{Horbert2015}.
We select this dataset as it contains sequences from challenging cluttered scenes with many objects (approximately 600 frames and 80 objects per sequence). 
This makes it more suitable for our envisioned application area of robotics than other datasets consisting mostly of single images.
Ground truth labels indicate the true objects for every 30\textsuperscript{th} frame.

We tuned our method and trained the proposal scoring SVM on the first sequence of the data set, and apply it to the remaining four sequences, labeled Kitchen A, B, C, and D, for testing.
For superpixel generation, we apply the SLIC algorithm~\cite{Achanta2012} with a target of 1000 superpixels with a compactness of 45.
Features for superpixels are computed as described in Section~\ref{sec:gestalt_principles_for_object_discovery}.
We set a self-link likelihood as $\log \alpha = 0$.
For the exponential decay function $f(d)=\exp(-d/a)$, we set $a = 0.05$.
For the base measure, we set $m_0 = \begin{bmatrix}1 & 1 & 1 \end{bmatrix}^T$ with a low confidence $\kappa_0 = 0.1$, and $S_0 = 10 \cdot \text{I}_{3\times 3}$ with $v_0 = 5$.

For each image, we draw $M=50$ samples of segmentations applying the ddCRP.
Samples from a burn-in period of 50 samples were first discarded to ensure the underlying Markov chain enters its stationary distribution.
We rank the proposals applying the score $s(o)$ or the likelihood-weighted score $s_w(o)$, and return up to 200 proposals with the highest score.
Before ranking we removed proposals larger than 10\% or smaller than 0.1\% of the image size.

We compare our method to the saliency-guided object candidates (SGO) of~\cite{Werner2015}, the objectness measure (OM) of~\cite{Alexe2012}, and the randomized Prim's algorithm (RP) of~\cite{Manen2013}.
SGO is a recent method that performs well on the KOD dataset.
The other two methods are representatives of the window-scoring (OM) and segment-grouping (RP) streams of object discovery methods.
We measure precision and recall in terms of the number of valid object proposals that have IoU $\geq$ 0.5 with the ground truth.
As OM outputs proposals as bounding boxes, we evaluate all methods with bounding boxes for a fair comparison.
We define the bounding box of a proposal as the smallest rectangle enclosing the whole proposal.

\begin{table}[t]
\centering
\caption{Area under curve (AUC) values for precision and recall averaged over all frames on the test data sequences labeled A through D, and averaged over all test sequences. ``Weighted'' refers to using the score $s_w(o)$, ``plain'' to using the score $s(o)$. Non-maxima suppression (NMS) was applied in all cases. The greatest values for each sequence are shown in a bold font.}
\label{tab:auc}
\begin{tabular}{@{}ccccccccccccccc@{}}
\toprule
                      & \multicolumn{2}{c}{Kitchen A} &  & \multicolumn{2}{c}{Kitchen B} &  & \multicolumn{2}{c}{Kitchen C} &  & \multicolumn{2}{c}{Kitchen D} & & \multicolumn{2}{c}{Average}\\ 
                      \cmidrule(lr){2-3} \cmidrule(lr){5-6} \cmidrule(lr){8-9} \cmidrule(l){11-12} \cmidrule(l){14-15} 
                      & Prec.          & Rec.         &  & Prec.          & Rec.         &  & Prec.          & Rec.         &  & Prec.          & Rec.          & & Prec. & Rec.       \\ \midrule
Ours (weighted)       & \textbf{19.4}  & \textbf{93.3}&  & 25.2           & \textbf{86.0}&  & 12.1           & \textbf{86.7}&  & 26.7           & 47.4          & & \textbf{20.8} & \textbf{78.3}  \\
Ours (plain)          & 16.8           & 83.1         &  & 22.2           & 79.1         &  & 11.8           & 85.3         &  & \textbf{27.9}  & 49.2          & & 19.7 & 74.2  \\
SGO~\cite{Werner2015} &  9.8           & 60.9         &  & \textbf{25.3}  & 85.5         &  &  9.6           & 81.6         &  & \textbf{27.9}  & \textbf{51.9} & & 18.2 & 70.0  \\ 
OM~\cite{Alexe2012}   & 11.5           & 45.7         &  & 14.7           & 44.4         &  & \textbf{18.1}  & 83.8         &  &  8.6           & 17.2          & & 13.2 & 47.8  \\ 
RP~\cite{Manen2013}   & 11.1           & 61.2         &  & 12.3           & 46.0         &  & 12.0           & 70.0         &  & 11.8           & 25.2          & & 11.8 & 50.6  \\ \bottomrule
\end{tabular}
\end{table}

\begin{figure}[t]
	\centering
	\includegraphics[width=\columnwidth]{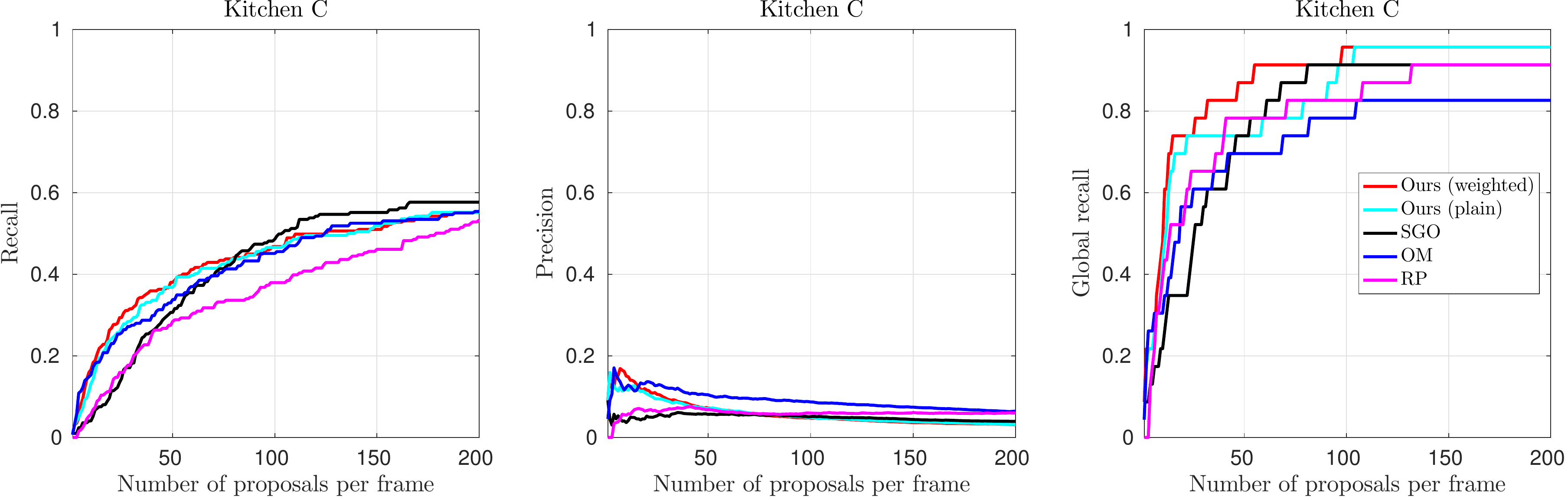}
	\caption{From left to right: recall, precision, and global recall (fraction of all objects in the sequence detected) averaged over all frames in the Kitchen C sequence. The results are shown as a function of the number of best-ranked proposals considered.}
	\label{fig:kitchen_c}
\end{figure}

The results are summarized in Table~\ref{tab:auc}.
As shown by the average column, the proposed method with likelihood weighting performs best both in terms of precision and recall.
With the plain scoring we still slightly outperform SGO, OM, and RP.
On individual sequences, we reach the performance of SGO on sequences B and D, while outperforming it on A and C.
OM has better precision and similar recall as our method and SGO on sequence C, but does not perform as well on other sequences.
On sequences A, B, and C, applying our likelihood-weighted proposal scoring improves performance compared to the plain scoring method.
Thus, the likelihood is useful for ranking proposals, providing complementary information not available with the plain score.

For sequence C, the recall, precision, and global recall (fraction of all objects in the sequence detected over all frames) as a function of the number of best-ranked proposals considered are shown in Fig.~\ref{fig:kitchen_c}.
We achieve higher precision and global recall than SGO for a low number of proposals ($<$ 50) per frame.
We achieve greater global recall than all the other methods, detecting a greater fraction of all objects over the whole sequence.

\begin{figure}[t]
	\centering
	\includegraphics[width=\columnwidth]{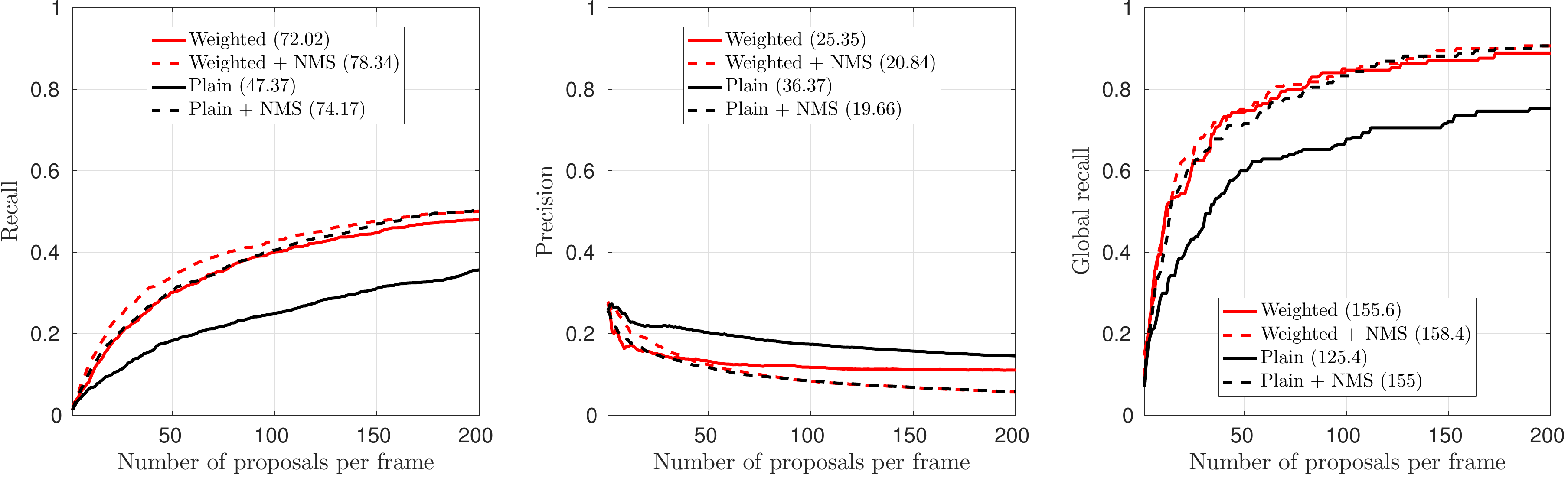}
	\caption{Evaluation of the ranking methods. Plain refers to the score $s(o)$, weighted is the likelihood weighted score $s_w(o)$, while NMS indicates applying non-maxima suppression (duplicate removal). The numbers in parenthesis show the AUC values for each curve.}
	\label{fig:ranking}
\end{figure}

Fig.~\ref{fig:ranking} shows the effect of ranking method on the performance of our method when averaging over all of the four sequences.
Applying likelihood-weighting together with non-maxima suppression (NMS) improves the results over applying the plain score.
Applying NMS decreases the reported precision, since it removes also good duplicates from the set of proposals.

\begin{figure}[h!t!]
	\centering
	\includegraphics[width=\columnwidth]{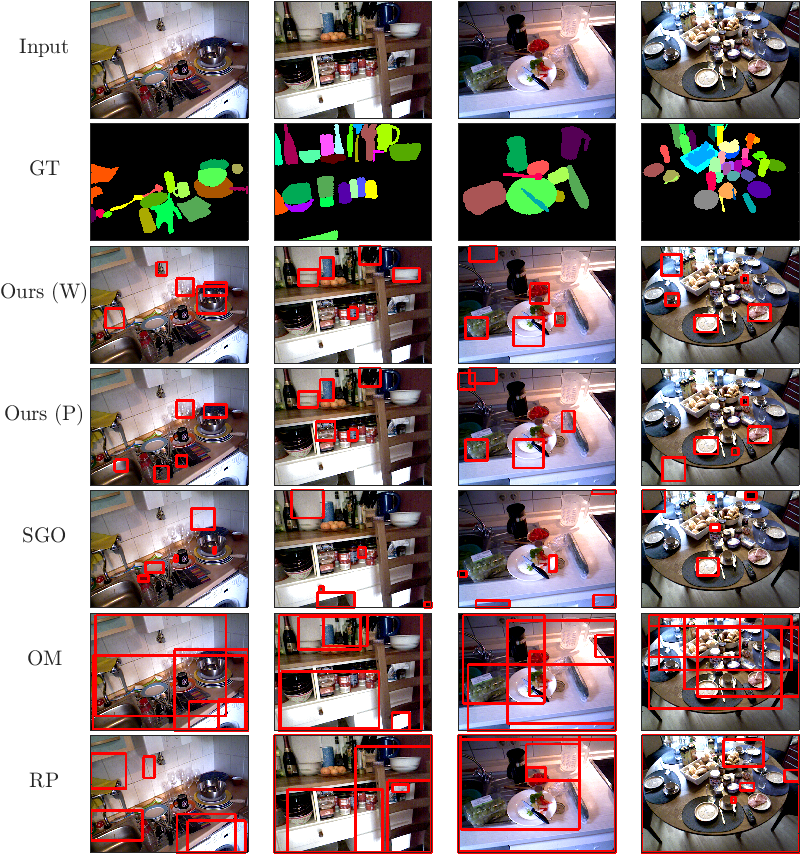}
	\caption{Bounding boxes for the top 5 object proposals. From top to bottom: input image, ground truth labels, ours (likelihood weighted), ours (plain score), SGO~\cite{Werner2015}, OM~\cite{Alexe2012}, and RP~\cite{Manen2013}. From left to right: one frame from sequence A, B, C, or D.}
	\label{fig:qualitative}
\end{figure}

Fig.~\ref{fig:qualitative} qualitatively compares the 5 best proposals from each of the methods.
OM and RP tend to produce large object proposals (last two rows).
The third and fourth row show the likelihood weighted and plain scoring, respectively.
Compared to plain scoring, likelihood weighting increases the rank of proposals that appear often in the ddCRP samples.
For example, in the last column, fourth row, the plain score gives a high rank for the patch of floor in the lower left corner and the patch of table covering in the lower middle part of the image.
These proposal rarely appear in the ddCRP samples.
With likelihood weighting (last column, third row), the often appearing proposals on the coffee cup in the middle left part and near the glass in the top left part of the image are preferred as they have a higher likelihood, as also seen from Fig.~\ref{fig:seg_example}. 


\section{Conclusion} 
\label{sec:conclusion}
We introduced object proposal generation via sampling from a distance dependent Chinese restaurant process posterior on image segmentations.
We further estimated a likelihood value for each of the proposals.
Our results show that the proposed method achieves state-of-the-art performance, and that the likelihood estimate helps improve performance.
Further uses for the likelihood estimates may be found, e.g., in robotics applications.
Other future work includes extending the method to RGB-D data, and an analysis of the parameter dependency.


\bibliographystyle{splncs03}
\bibliography{ref}

\end{document}